\documentclass[journal,twoside,web]{ieeecolor}
\usepackage{amsmath}
\usepackage{amssymb}
\usepackage{bm}
\usepackage{mathtools}
\usepackage{verbatim}
\usepackage{tablefootnote}

\usepackage{amsmath,amsfonts}
\usepackage{algorithmicx}  
\usepackage[ruled,linesnumbered]{algorithm2e}
\usepackage{threeparttable}

\usepackage{bbold}
\usepackage{bbm}
\usepackage{booktabs} %
\usepackage{array}  %
\usepackage[tight,normalsize,sf,SF]{subfigure}
\usepackage{multirow}
\usepackage{tabularx}
\usepackage{url}
\usepackage{color}
\usepackage{float}
\usepackage{gensymb}
\usepackage{cite}

\usepackage{tabu}
\usepackage{dcolumn}

\usepackage{graphicx}

\usepackage[pagebackref=true,breaklinks=true,letterpaper=true,colorlinks,bookmarks=false]{hyperref}

\usepackage{generic}
\usepackage{cite}
\usepackage{amsmath,amssymb,amsfonts}
\usepackage{graphicx}
\usepackage{textcomp}

\def\BibTeX{{\rm B\kern-.05em{\sc i\kern-.025em b}\kern-.08em
    T\kern-.1667em\lower.7ex\hbox{E}\kern-.125emX}}
\begin{document}
\title{Prior Normality Prompt Transformer for Multi-class Industrial Image Anomaly Detection}
\author{Haiming Yao, Yunkang Cao, Wei Luo, Weihang Zhang, \\ Wenyong Yu, \IEEEmembership{Senior Member IEEE}, and Weiming Shen, \IEEEmembership{Fellow IEEE}
\thanks{Manuscript received XX XX, 20XX; revised XX XX, 20XX. This study was supported in part by the National Natural
Science Foundation of China (Grant No. 52375494), and in part by the Beijing Institute of Technology Research Fund Program for Young Scholars (Grant No. 6120220236). (Corresponding author: Wenyong Yu.)}
\thanks{Haiming Yao and Wei Luo are with the State Key Laboratory of Precision Measurement Technology and Instruments, Department of Precision Instrument, Tsinghua University, Beijing 100084, China. (e-mails: yhm22@mails.tsinghua.edu.cn; luow23@@mails.tsinghua.edu.cn).}
\thanks{Yunkang Cao, Wenyong Yu, and Weiming Shen are with the State Key Laboratory of Digital Manufacturing Equipment and Technology, School of Mechanical Science and Engineering,
Huazhong University of Science and Technology, Wuhan 430074, China (e-mails: cyk$\_$hust@hust.edu.cn; ywy@hust.edu.cn; wshen@ieee.org).}
\thanks{Weihang Zhang is with the School of Medical Technology, Beijing Institute of Technology, Beijing, 100081, China (e-mail: zhangweihang@bit.edu.cn).}}

\maketitle

\begin{abstract}


Image anomaly detection plays a pivotal role in industrial inspection. Traditional approaches often demand distinct models for specific categories, resulting in substantial deployment costs. This raises concerns about multi-class anomaly detection, where a unified model is developed for multiple classes. However, applying conventional methods, particularly reconstruction-based models, directly to multi-class scenarios encounters challenges such as identical shortcut learning, hindering effective discrimination between normal and abnormal instances. To tackle this issue, our study introduces the Prior Normality Prompt Transformer (PNPT) method for multi-class image anomaly detection. PNPT strategically incorporates normal semantics prompting to mitigate the "identical mapping" problem. This entails integrating a prior normality prompt into the reconstruction process, yielding a dual-stream model. This innovative architecture combines normal prior semantics with abnormal samples, enabling dual-stream reconstruction grounded in both prior knowledge and intrinsic sample characteristics. PNPT comprises four essential modules: Class-Specific Normality Prompting Pool (CS-NPP), Hierarchical Patch Embedding (HPE), Semantic Alignment Coupling Encoding (SACE), and Contextual Semantic Conditional Decoding (CSCD). Experimental validation on diverse benchmark datasets and real-world industrial applications highlights PNPT's superior performance in multi-class industrial anomaly detection.

\end{abstract}


\section{Introduction}
\label{sec:introduction}
\IEEEPARstart{T}{he} realm of anomaly detection (AD) in industrial image analysis has gained considerable attention, driven by the increasing demand for automation in smart manufacturing. This application proves versatile, addressing diverse needs such as railway track inspection~\cite{r1}, textured surface defect detection~\cite{r8}, and the examination of various product surfaces~\cite{r3}. Current research predominantly favors unsupervised learning anomaly detection due to the limited availability of abnormal samples.

\begin{figure*}[t]
\centerline{\includegraphics[width=170mm]{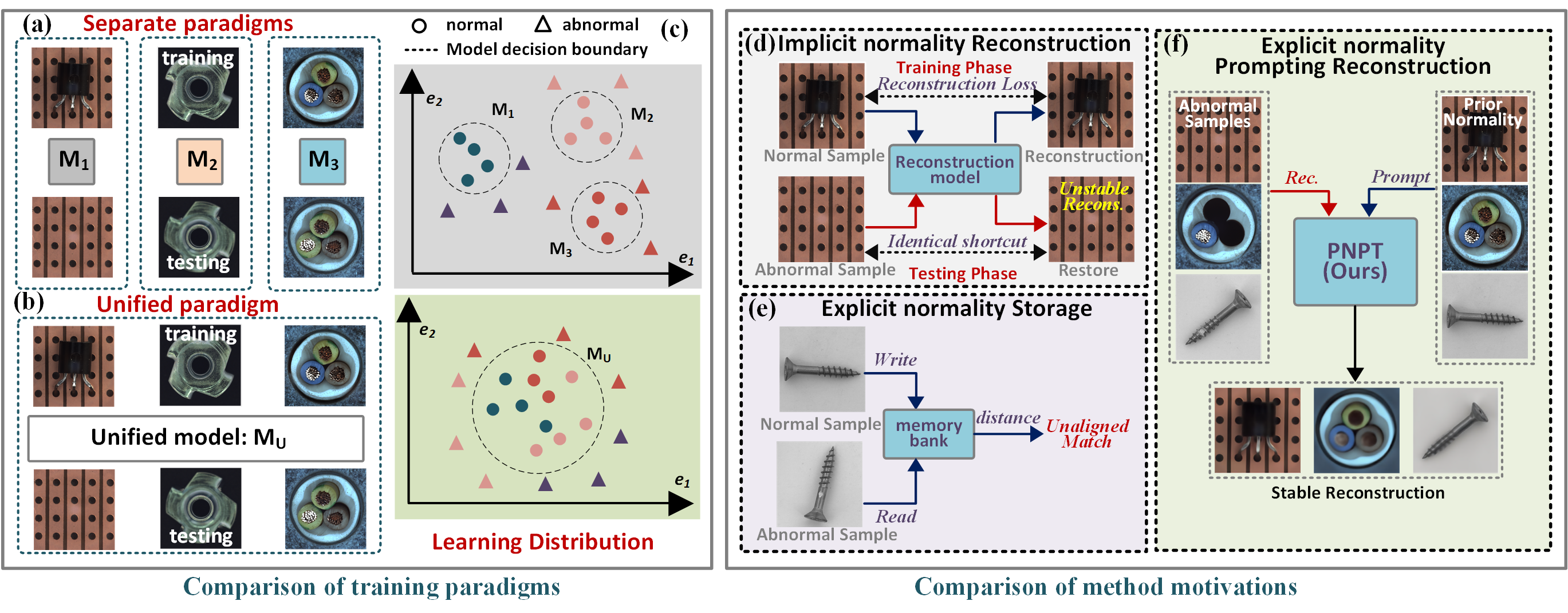}}
\caption[width=170mm]{Left:\textbf{ Comparison between the separate single-class training and unified multi-class training paradigms}. (a) The separate training mode needs to assign distinct weights specific to individual categories. (b) The unified model requires only a shared weight to simultaneously execute the detection task across multiple categories. (c) The learning distributions for two training paradigms.
Right: \textbf{Comparison of method motivations}. (d) Implicit learning methods, characterized by the implicit introduction of normal semantics during training that is subsequently omitted during testing, lead to unstable reconstruction prone to identical mapping. (e) Explicit learning methods directly and explicitly compare the samples with normal semantic templates in the memory bank, thereby being influenced by misalignment factors. (f) In contrast, the proposed normality prompting framework introduces normal semantics as prompt information for stable reconstruction.}
\label{fig1}
\end{figure*}
In Fig.~\ref{fig1} (a), when simultaneous detection of multiple categories is necessary, existing unsupervised methods often adopt a separate training mode. Here, each category undergoes individual training, and category-specific weights are retained. However, this one-class-one-weight approach becomes memory-intensive with numerous categories to detect. As shown in Fig.~\ref{fig1} (b), this study explores a more efficient and universally applicable paradigm, \textit{i.e.}, multi-class anomaly detection. This involves employing a shared unified weight for comprehensive detection across various categories. Nevertheless, this task poses inherent challenges. As depicted in Fig.~\ref{fig1} (c), in the individual training paradigm, each model establishes a distinct normal decision boundary for every category. In contrast, within the unified multi-class training paradigm, it is significantly harder to establish the boundary because of intricate distributions across multiple categories.

Various methods are proposed to describe the normal feature distribution through the inherent representation capabilities of deep neural networks. A commonly used reconstruction-based scheme~\cite{r4} involves reconstructing abnormal defect samples into templates grounded in normal semantics, as illustrated in Fig.~\ref{fig1} (d). The dissimilarity between the reconstructed templates and the abnormal samples forms the basis for localizing and detecting abnormal defects. However, this method introduces normal prior information implicitly during training, neglecting its utility during testing. Consequently, there is a tendency to resort to a shortcut reconstruction process of "identical mapping," where the model reproduces an identical copy of the input without considering its semantics. This challenge is amplified, especially in the context of the aforementioned multi-class training mode, where the model must learn intricate distributions. To explicitly incorporate normal information during testing, alternative approaches involve storing normal semantics for subsequent retrieval~\cite{r5}, as depicted in Fig.~\ref{fig1} (e). However, these methods often incur significant storage and computational overhead in multi-class training scenarios. Moreover, when the memory bank size is constrained, the diversity of normal semantics is inevitably limited. Detection performance may also degrade due to factors such as misalignment, making it challenging to retrieve analogous normal semantic templates for test images.

In recent times, prompting~\cite{r7} has gained significant prominence in language-visual multi-modal systems, such as visual grounding~\cite{r29}. In these frameworks, models are dynamically regulated for specified tasks based on provided prompt information. Motivated by this, this study introduces a novel concept termed “prior normality prompt" as depicted in Fig.~\ref{fig1} (f). This involves incorporating prior normal semantics into the reconstruction-based AD model, employing it as prompt information to guide the model in reconstructing abnormal samples into templates with normal semantics. This explicit integration facilitates a more stable reconstruction process, mitigating the “identity mapping” reconstruction and proving highly advantageous for anomaly detection, particularly in the context of multi-class training scenarios. Furthermore, this approach circumvents the computationally intensive memory bank-based retrieval process and mitigates the misalignment factors that can arise from limited samples in the memory bank, tackling the existing challenges effectively. 

To dynamically regulate the reconstruction process with the “prior normality prompt", this study presents a framework involving a Prior Normality Prompt Transformer (PNPT) for multi-class industrial image AD. Within this framework, this study extends beyond the conventional single-stream approach of input sample and introduces an additional prompting stream grounded in prior normality. This results in the construction of a dual-stream framework, which enables robust reconstruction through semantic alignment of prior normality prompting and sample self-attribute. Our core contributions can be summarized as follows.
\begin{enumerate}
    \item This study introduces a novel PNPT method for multi-class industrial image AD, which employs a dual-stream hybrid paradigm integrating explicit and implicit learning methods

    \item PNPT incorporates four key designs: the CS-NPP, HPE, SACE, and CSCD modules. These modules operate on the semantic alignment between normal prompting and sample self-attributes, to ensure stable reconstruction and avoid the "identical mapping" issue.

    \item Extensive experiments on public datasets validate the superior performance of PNPT in multi-class industrial image anomaly detection. Specifically, in the multi-class scenario, it achieves image-level and pixel-level AUROC scores of 98.3 and 98.1, respectively, on the representative MVTec AD dataset. Additionally, it also demonstrates superior performance on the MVTec LOCO and BTAD datasets, and we have further validated its effectiveness in practical industrial experiments.
\end{enumerate}

\begin{figure*}[t]
\centerline{\includegraphics[width=165mm]{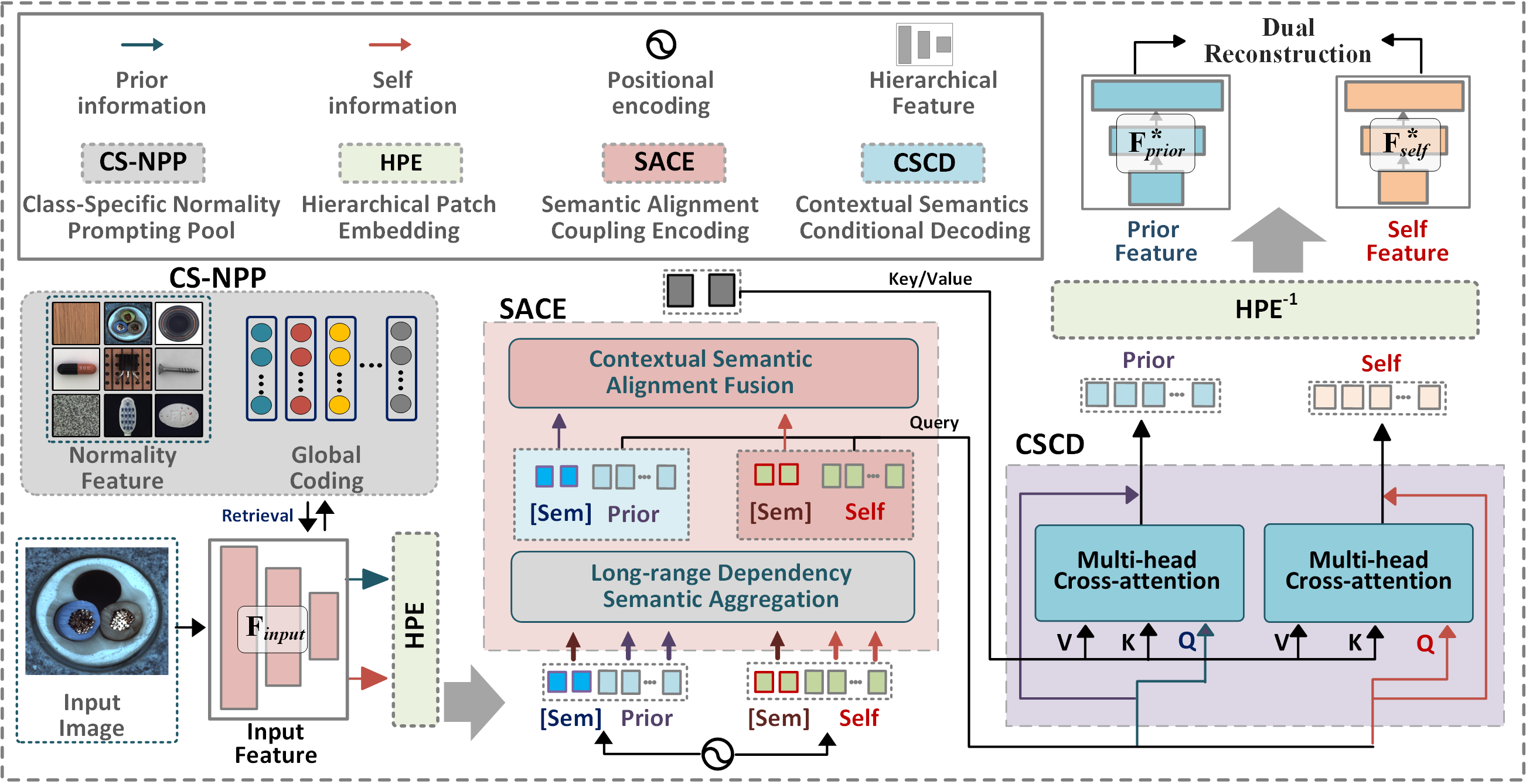}}
\caption[width=165mm]{
\textbf{PNPT Framework}: PNPT employs a dual information flow structure by leveraging the CS-NPP to extract category-specific normal prompt information for the input image. This extraction process facilitates the construction of dual input features encompassing both the normal prior and the sample itself. The dual input features are then transformed into token sequences through HPE. Then, the sequences are encoded via SACE. In SACE, semantic tokens $[\mathbf{Sem}]$ are incorporated into the two branch patch tokens, and the joint sequences undergo long-distance semantic dependency aggregation for acquiring the high-level semantics and the encoded patch tokens. These high-level semantics then pass through the Contextual Semantic Alignment Fusion module for semantics alignment. The decoding occurs in CSCD, using two branch patch tokens as queries and aligned semantics tokens as keys and values. The reconstruction features are obtained via the reverse process of HPE(denoted as HPE$^{-1}$).
}
\label{fig2}
\end{figure*}
\section{Related work}
\label{sec:guidelines}
This study categorizes existing industrial image AD methods into two primary categories: implicit and explicit normality learning methods, which are introduced below.

\subsection{Implicit normality learning approaches}

The implicit learning method involves incorporating normal samples during training, implicitly integrating their semantic information into the weights of models. Normal samples are no longer utilized during testing. Representative schemes falling under this category include reconstruction-based~\cite{r4} and regression-based~\cite{r9} methods, where the model is trained to either reconstruct or regress the features of normal samples. Abnormal samples typically exhibit larger reconstruction or regression errors during the testing phase. The reconstruction-based method encompasses both image reconstruction~\cite{r4}~\cite{r39} and feature reconstruction~\cite{r10,r37, r38}, with autoencoders serving as a representative architecture.

In the context of image reconstruction, previous studies have introduced structural loss-driven autoencoders~\cite{r8} and utilized inpainting-based reconstruction methods~\cite{r24}. Feature reconstruction, on the other hand, involves techniques like DFR~\cite{r10}, which employs multi-level pre-trained features as reconstruction objectives. Additionally, in ~\cite{r4}~\cite{r25}~\cite{r36}, the concept of synthetic anomalies was introduced. The regression-based method is exemplified by knowledge distillation, initially employed in the context of AD within the teacher-student framework in ~\cite{r9}. Over time, this approach has seen numerous enhancements~~\cite{r16,r26,r25,r27}, presenting different strategies for alleviating the shortcut learning problem.

However, a notable limitation of this category of methods pertains to the shortcut learning phenomenon of ”identical mapping”~\cite{r13, r25} observed during the training process, especially in multi-class scenarios~\cite{r14}. In such instances, the model tends to learn direct input copying without fully incorporating semantic information. As a result, both abnormal and normal samples exhibit comparable reconstruction or regression errors, leading to a deterioration in detection performance.

\subsection{Explicit normality learning approaches}
The explicit learning method involves a direct comparison between the attributes of the test sample and those of normal samples, serving as the criterion for abnormality during testing. This approach was initially introduced in SPADE~\cite{r18}, which directly retrieves the nearest item from a normal sample database and uses their distances as anomaly scores. Subsequently, Padim~\cite{r19} leverages the Gaussian distribution of normal samples, while statistical features are employed for anomaly scoring. An important development in this category is Patchcore~\cite{r5}, which utilizes a coreset-sampling memory bank to alleviate the storage requirements. This method has seen further extensions in subsequent works~\cite{r28, r20}.

This category of methods generally exhibits more robust detection performance. However, there are notable drawbacks. Firstly, the storage of a considerable volume of normal sample attributes can be memory-intensive. Secondly, the diversity of normal attributes in the memory bank significantly impacts detection performance. For instance, situations may arise where no normal samples align with the test sample in the memory bank, leading to inaccurate results in comparisons.

The PNPT method is innovative for integrating both learning methods, explicitly incorporating the prior properties of normal samples in the implicit reconstruction process for robust multi-class industrial image AD.

\section{PNPT methodology}
\subsection{Model overall structure}

As mentioned before, the implicit normality learning method omits the direct involvement of normal semantics in the model inference process, leading to the phenomenon of shortcut learning through "identical mapping." Conversely, the explicit learning method faces constraints related to computational overhead and non-alignment factors. The PNPT framework introduced in this study represents an innovative integration of these two learning approaches. This integration enhances the implicit reconstruction-based method by incorporating explicit normality prompt information to integrate prior semantics with the intrinsic semantics of the samples.

The overall architecture of PNPT is illustrated in Fig.~\ref{fig2}. The proposed framework consists of four main novel components: Class-Specific Normality Prompting Pool (CS-NPP), Hierarchical Patch Embedding (HPE), Semantic Alignment Coupling Encoding (SACE), and Contextual Semantic Conditional Decoding (CSCD).

First, unlike existing methods that only use the single input image, our PNPT adopts a dual-stream input paradigm within the CS-NPP, where the normality prior is introduced in addition to the input sample, which serves as prompt information for the reconstruction of the input sample. Consequently, this yields a collection of the sample itself and prior normality dual input features $\left \{ \mathbf{F}_{prior},\mathbf{F}_{self} \right \}$. Next, we employ the proposed HPE to establish the mapping transformation between features and sequences, converting the two features into sequences of feature patch tokens $\left \{ \mathbf{E}_{prior},\mathbf{E}_{self} \right \}$. Subsequently, $\left \{ \mathbf{E}_{prior},\mathbf{E}_{self} \right \}$ undergoes encoding via SACE and decoding through CSCD to yield the output sequence $\mathbf{E}_{prior}^\ast,\mathbf{E}_{self}^\ast$. Ultimately, the output sequences are mapped back into the reconstructed features $\big \{ \mathbf{F}_{prior}^{\ast},\mathbf{F}_{self}^{\ast} \big \}$ through the inverse process of HPE.

\begin{figure}[t]
\centerline{\includegraphics[width=80mm]{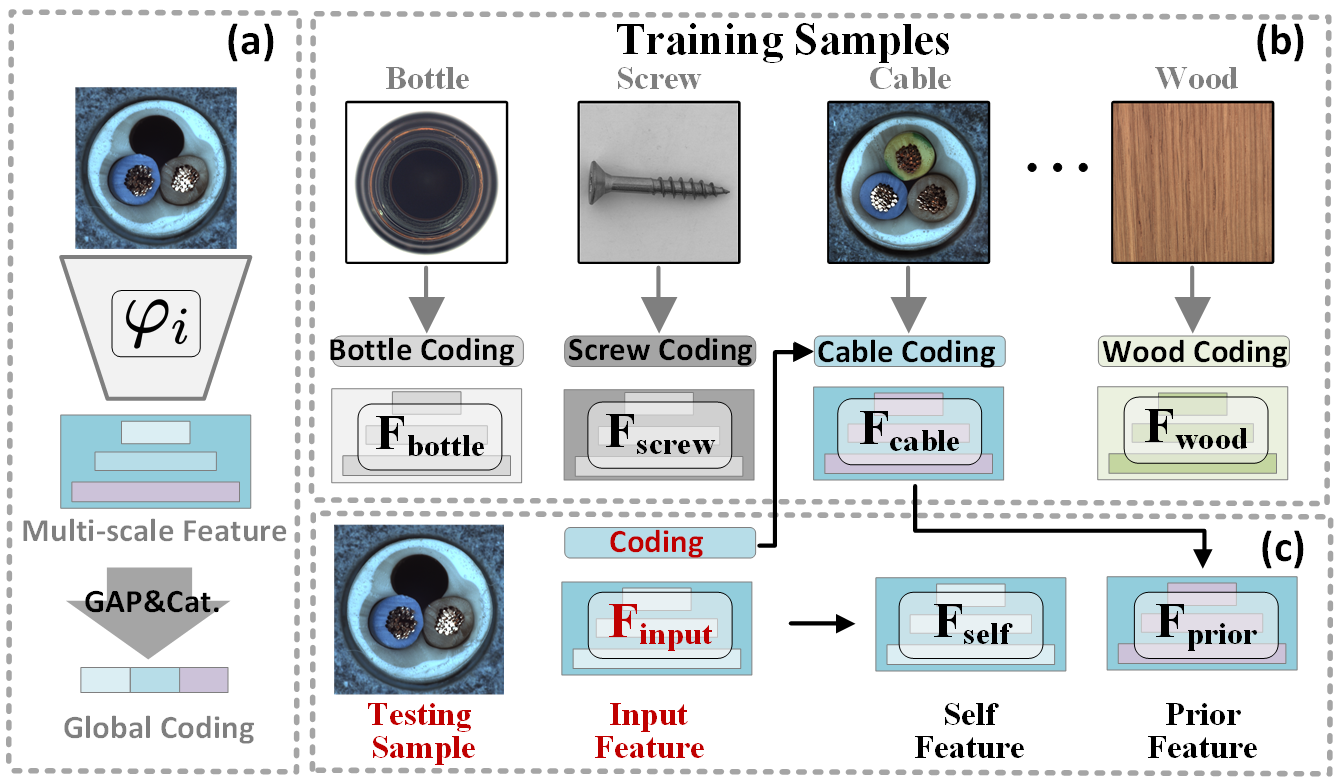}}
\caption[width=80mm]{
CS-NPP diagram. (a) Multi-scale feature and global coding acquisition process. (b) Class-specific normality feature and query formation. (c) dual feature input construction. GAP and Cat. represent global average pooling and concatenation.}
\label{fig3}
\end{figure}

\subsection{Class-Specific Normality Prompting Pool (CS-NPP)}

Distinguishing itself from conventional implicit approaches, PNPT incorporates an additional branch comprising prior knowledge of normal patterns. This auxiliary branch serves as prompting information during the reconstruction process of input samples, providing a representation of the typical normal pattern. To this end, this study introduces the CS-NPP module.

As shown in Fig.~\ref{fig3} (a), the input sample is fed into a convolutional neural network (CNN) encoder $\bm{\varphi_e}$ pre-trained on the ImageNet dataset to extract multi-scale features:
\begin{equation}
\mathbf{F}_{input}=\left \{ \mathbf{F}_{input}^{(i)}\in \mathbb{R}^{C^{i}\times H^{i}\times W^{i}}, i=1,.., \mathcal{H} \right \} 
\end{equation}
where the hierarchical features are captured from the output of the final layers of the first $\mathcal{H}=3$ convolutional modules of $\bm{\varphi_e}$. Subsequently, each scale of feature undergoes global average pooling, followed by the concatenation of different scales. This results in a global coding.

In multi-class scenarios as shown in Fig.~\ref{fig3} (b), we extract multi-scale features and global codings from all training normal samples within each class. Afterward, we average all features and codings corresponding to each category to obtain the class-specific normality feature and coding and thus establish the CS-NPP.

After the CS-NPP is established, we can generate a dual-stream input illustrated in Fig.~\ref{fig3} (c). When provided with an input sample, we designate its input features as self-features $\mathbf{F}_{self}$ and leverage its global coding to retrieve the nearest class coding within the CS-NPP. Subsequently, the retrieved class features serve as the class-specific prior normal features $\mathbf{F}_{prior}$ for the given sample. Therefore, the above two features constitute the dual-stream input $\left \{ \mathbf{F}_{prior},\mathbf{F}_{self} \right \}$.

It is noteworthy that our CS-NPP implements category-level retrieval, leading to a significant reduction in computational complexity and storage overhead compared to the patch-level retrieval utilized in the existing explicit method PatchCore~\cite{r5}. Table~\ref{table_retrieval} shows the comparison of retrieval times between Patchcore~\cite{r5} and CS-NPP on the MVTec AD~\cite{r3} dataset.

\begin{table}
\caption{Comparison of retrieval times}
\setlength{\tabcolsep}{3pt}
\centering
\begin{tabular}{p{80mm}}
${\includegraphics[width=80mm]{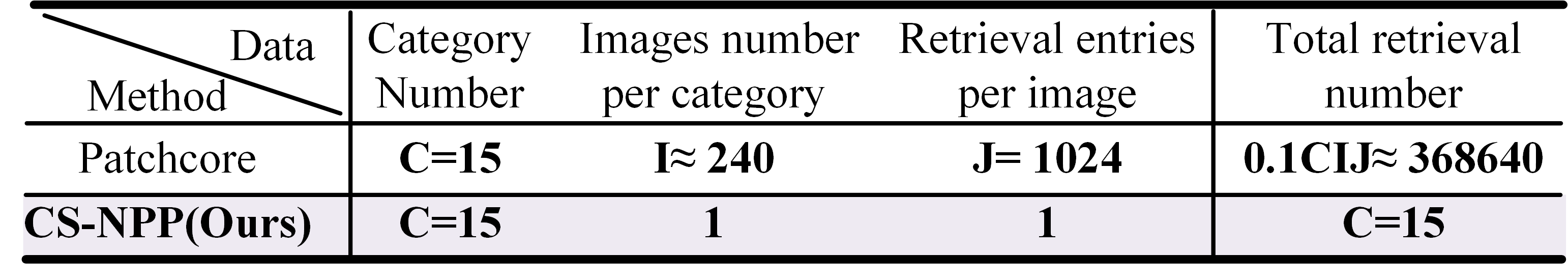}}$
\end{tabular}
\label{table_retrieval}
\end{table}

\begin{figure}[t]
\centerline{\includegraphics[width=80mm]{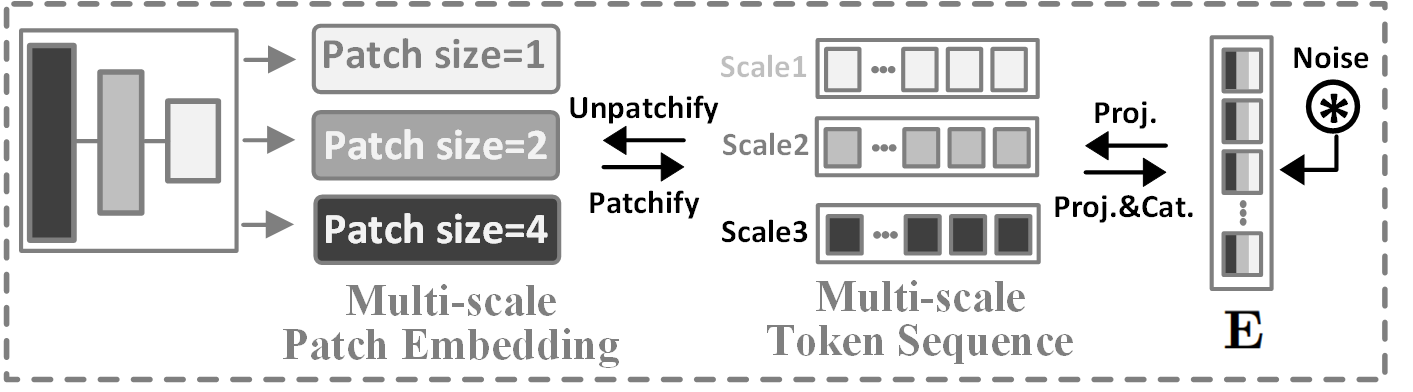}}
\caption[width=80mm]{
HPE schematics. The Proj. and Cat. represent projection and concatenation.}
\label{fig4}
\end{figure}
\subsection{Hierarchical Patch Embedding (HPE)}

To facilitate bidirectional mapping transformation between multi-scale 2D features and sequences, we introduced the Hierarchical Patch Embedding (HPE) module as illustrated in Fig.~\ref{fig4}.

For the forward process, to handle the 2D feature maps, we reshape the feature map into a sequence of flattened 2D patch tokens by employing a multiple-increasing multi-scale patch size strategy. This approach ensures uniform sequence length across features at different scales. Subsequently, the sequences from the three scales are mapped to hidden dimensions and then concatenated to yield the final feature patch embedding sequence $\mathbf{E} \in \mathbb{R}^{L\times C}$. Finally, the Gaussian noise is added to prevent over-fitting~\cite{r14}.

In the inverse process, a symmetrical operation to the aforementioned procedure is employed to transform the sequence back into multi-scale features.

\subsection{Semantic Alignment Coupling Encoding (SACE)}

The Semantic Alignment Coupling Encoding (SACE) module principally involves two sub-processes: long-distance semantic dependency aggregation based on self-attention and contextual semantic alignment fusion based on second-order cross-attention. These procedures are implemented to attain contextual semantic extraction and alignment between the sample itself and the prior information. The detailed structure of SACE is shown in Fig.~\ref{fig5}.

\subsubsection{Long-Distance Dependency Semantic Aggregation}
\begin{figure}[t]
\centerline{\includegraphics[width=80mm]{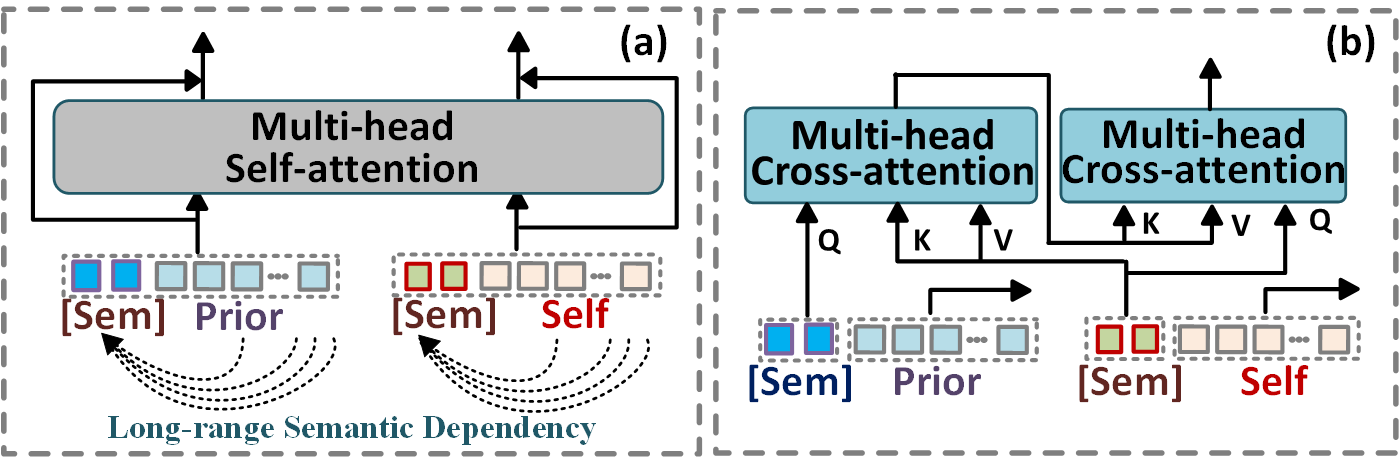}}
\caption[width=80mm]{SACE diagram. (a) Long-Distance Dependency Semantic Aggregation. (b) Contextual Semantic Alignment Fusion.}
\label{fig5}
\end{figure}

As illustrated in Fig.~\ref{fig5} (a), a group of semantic tokens $\mathbf{E}_{\mathcal{S}} \in\mathbb{R} ^{N\times C}$  from the preceding CSCD or the initialization is appended to two distinct feature patch token sequences $\mathbf{E}_{prior},\mathbf{E}_{self}$:
\begin{equation}
\mathbf{E}_{\mathcal{X+S}}=\left [ \underbrace{\mathbf{E}_{\mathcal{X},1},\mathbf{E}_{\mathcal{X},2},...,\mathbf{E}_{\mathcal{X},L}}_{\mathrm{feature\;tokens} } ,\underbrace{\mathbf{E}_{\mathcal{S},1},\mathbf{E}_{\mathcal{S},N}}_{\mathrm{semantic\;tokens}} \right ] 
\end{equation}
where the $\mathbf{E}_{\mathcal{X+S}}\in \mathbb{R}^{(L+N)\times C}$ represent the joint token sequences, and $\mathcal{X}\in \left \{\mathrm{prior} ,\mathrm{self } \right \}$, respectively. Subsequently, the self-attention mechanism is applied to the joint token sequences, and the procedure in the $l$-th SACE can be described as follows:
\begin{equation}
\begin{aligned}
\mathcal{Z}^{(l)}_{\mathcal{X+S}}&=\mathrm{LN}\Big( \mathrm{SAM} (\mathbf{E}_{\mathcal{X+S}}^{(l-1)},\mathbf{E}_{\mathcal{X+S}}^{(l-1)},\mathbf{E}_{\mathcal{X+S}}^{(l-1)})+\mathbf{E}_{\mathcal{X+S}}^{(l-1)} \Big )\\
\mathbf{R}_{\mathcal{X+S}}^{(l)}&=\mathrm{LN}\Big( \mathrm{FP} (\mathcal{Z}^{(l)}_{\mathcal{X+S}})+\mathcal{Z}^{(l)}_{\mathcal{X+S}}\Big ) ,\mathcal{X}\in \left \{\mathrm{prior} ,\mathrm{self } \right \}
\end{aligned}
\end{equation}
where the $\mathrm{SAM} (\mathbf{Q},\mathbf{K},\mathbf{V} )$ denotes the self-attention mechanism, encompassing the query $\mathbf{Q}$, key $\mathbf{K}$, and value $\mathbf{V}$ embeddings, $\mathcal{Z}$ is the hidden representation, and $\mathbf{R}$ is the output. LN represents layer normalization, and FP denotes a forward propagation layer. 

It is noteworthy that, owing to the attention mechanism, mutual correspondences can be freely established between each pair of tokens within the joint entity $\mathbf{E}_{\mathcal{X+S}}$. Consequently, the output state of the semantics tokens typically evolves into a contextually enriched and long-distance semantics aggregated representation that encompasses all feature patch tokens. Furthermore, each feature patch token can also undergo encoding by interacting with other feature tokens.

Following that, the two patch token sub-sequences $\mathbf{R}_{\mathcal{X}}^{(l)}$ in encoded joint sequences $\mathbf{R}_{\mathcal{X+S}}^{(l)}$ of sample self-attribute and prior normality branches are then utilized as queries for CSCD. However, the two semantic token sub-sequences $\mathbf{R}_{\mathcal{S}}^{(l)}$ will be directed to the subsequent stage of contextual semantic alignment fusion.

\subsubsection{Contextual Semantic Alignment Fusion}
After acquiring the semantic token sequences $\mathbf{R}_{\mathcal{S}(\mathrm{s})}^{(l)}$ and $\mathbf{R}_{\mathcal{S}(\mathrm{p})}^{(l)}$ of both the sample itself and the prior normality branch, this study introduces a subsequent step for semantic alignment fusion of two branches. As illustrated in Fig.~\ref{fig5} (b), this involves employing the second-order cross-attention mechanism ($\mathrm{CAM} (\mathbf{Q},\mathbf{K},\mathbf{V} )$) defined as follows to align the two modality contextual semantics:
\begin{equation}
\begin{aligned}
\mathbf{R}^{(l)'}_{\mathcal{S}(\mathrm{p})}&=\mathrm{LN}\Big( \mathrm{CAM} (\mathbf{R}_{\mathcal{S}(\mathrm{p})}^{(l)},\mathbf{R}_{\mathcal{S}(\mathrm{s})}^{(l)},\mathbf{R}_{\mathcal{S}(\mathrm{s})}^{(l)})+\mathbf{R}_{\mathcal{S}(\mathrm{p})}^{(l)}\Big )\\
\mathcal{Z}^{(l)'}_{\mathcal{S}}&=\mathrm{LN}\Big( \mathrm{CAM} (\mathbf{R}_{\mathcal{S}(\mathrm{s})}^{(l)},\mathbf{R}^{(l)'}_{\mathcal{S}(\mathrm{p})},\mathbf{R}^{(l)'}_{\mathcal{S}(\mathrm{p})})+\mathbf{R}_{\mathcal{S}(\mathrm{s})}^{(l)}\Big ) \\
\mathbf{E}^{(l)}_{\mathcal{S}}&=\mathrm{LN}\Big( \mathrm{FP} (\mathcal{Z}^{(l)'}_{\mathcal{S}})+\mathcal{Z}^{(l)'}_{\mathcal{S}}\Big ) 
\end{aligned}
\end{equation}

In this second-order strategy, the query embedding of the first-order CAM is derived from the prior branch, while the key and value embeddings are sourced from the sample branch. Consequently, the output of this CAM module consists of sample-attribute-coupled normal semantics. Subsequently, the coupled normal semantics function as both key and value in the second-order CAM, with the sample branch serving as the query. This facilitates the second-order CAM in achieving a fused representation that aligns the sample's self-attribute with the coupled normal semantics. 

Thus, the high-level semantics of the sample self-attribute are comprehensively integrated with those of the prior normality. Ultimately, the fused high-level semantic token sequence is obtained by executing FP operations, ultimately achieving the final contextual semantics $\mathbf{E}^{(l)}_{\mathcal{S}}$.

\subsection{Contextual Semantics Conditional Decoding (CSCD)}
In the SACE, this study encodes the two patch token sequences $\mathbf{R}_{\mathcal{X}}^{(l)},\mathcal{X}\in \left \{\mathrm{prior},\mathrm{self } \right \}$ from sample self-attribute and the prior normality branch, and obtain the fused contextual semantic token sequence $\mathbf{E}_{\mathcal{S}}^{(l)}$. Additionally, the CSCD module is introduced to enable the encoded token sequences to absorb the fused contextual semantics for decoding:
\begin{equation}
\begin{aligned}
\hat{\mathcal{Z}}^{(l)}_{\mathcal{X}}&=\mathrm{LN}\Big( \mathrm{CAM} (\mathbf{R}_{\mathcal{X}}^{(l)},\mathbf{E}_{\mathcal{S}}^{(l)},\mathbf{E}_{\mathcal{S}}^{(l)})+\mathbf{R}_{\mathcal{X}}^{(l)}\Big )\\
\mathbf{E}_{\mathcal{X}}^{(l)}&=\mathrm{LN}\Big( \mathrm{FP} (\hat{\mathcal{Z}}^{(l)}_{\mathcal{X}})+\hat{\mathcal{Z}}^{(l)}_{\mathcal{X}}\Big ) ,\mathcal{X}\in \left \{\mathrm{prior} ,\mathrm{self } \right \}
\end{aligned}
\end{equation}

Through CSCD, integrated contextual semantics, encompassing both the sample's intrinsic attributes and prior normality, are introduced into the patch tokens of the two branches. Subsequently, each branch undergoes decoding and reconstruction processes based on integrated semantics. The resulting reconstruction outcomes capture the distinctive properties of the sample and prior normality, effectively preventing the occurrence of "identical mapping" and the generation of duplicates identical to the input.

The final output $\mathbf{E}_{\mathcal{X}}^{(M)}, \mathcal{X}\in \left \{\mathrm{prior},\mathrm{self } \right \}$, obtained after $M$ iterations of the SACE and CSCD processes, are fed into the inverse process of HPE to generate the reconstructed features $\big \{ \mathbf{F}_{prior}^{\ast},\mathbf{F}_{self}^{\ast} \big \}$.

\subsection{Training and Inference}
\subsubsection{Training loss}
As PNPT operates as a reconstruction-based system, a hierarchical feature reconstruction approach is employed as the primary loss function. This involves reconstruction by normality prior $\mathbf{F}_{prior}$ and self-reconstruction for $\mathbf{F}_{self}$:

\begin{equation}
\ell_{rec}= \ell_{\mathcal{R}}(\mathbf{F}_{prior})+\ell_{\mathcal{R}}(\mathbf{F}_{self})
\end{equation}
where the $\ell_{\mathcal{R}}$ is the cosine similarity loss of the feature vectors:

\begin{equation}
\ell_{\mathcal{R}}(\mathbf{F}_{\gamma })= \sum_{i=1}^{\mathcal{H} }\left (  1- \frac{\mathrm{vec} (\mathbf{F}_{input}^{(i)})\cdot \mathrm{vec}(\mathbf{F}_{\gamma }^{(i)*})}{\left \| \mathrm{vec}(\mathbf{F}_{input}^{(i)})\right \|_2 \cdot \left \| \mathrm{vec}(\mathbf{F}_{\gamma }^{(i)*}) \right \|_2 } \right )
\end{equation}
where $\mathrm{vec}(\cdot)$ is a vectorization function that converts a tensor into a one-dimensional vector. After the loss is calculated, PNPT adopts an end-to-end training strategy to optimize the entire model.

\subsubsection{Anomaly Scoring}
During the testing phase, the reconstruction errors are employed as the anomaly score:
\begin{equation}
AS_{rec}=\sum_{i=1}^{\mathcal{H} }\Phi \left(  1- \frac{\mathbf{F}_{input}^{(i)}\cdot \mathbf{F}_{fuse}^{(i)*}}{\left \| \mathbf{F}_{input}^{(i)} \right \|_2 \cdot \left \| \mathbf{F}_{fuse}^{(i)*} \right \|_2 } \right)
\end{equation}
where the $\mathbf{F}_{fuse}^{*}$ is the weighted sum of $\mathbf{F}_{prior}^{*}$ and $\mathbf{F}_{self}^{*}$, $\Phi(\cdot)$ denotes the Interpolation operation. 

The overall workflow of PNPT is shown in Algorithm 1.

\begin{algorithm}[t]
\caption{PNPT Framework}
 \SetKwInOut{Input}{input}
\SetKwInOut{KwOut}{Modules}
\KwIn{Input Feature $\mathbf{F}_{input}$}
 \KwOut{\textbf{CS-NPP}, \textbf{HPE}, \textbf{SACE}, \textbf{CSCD}}
\KwResult{Dual reconstructed feature $\mathbf{F}_{prior}^{*}$ and $\mathbf{F}_{self}^{*}$}

\emph{Step1}.Construct dual-stream input: 
$\mathbf{F}_{prior},\mathbf{F}_{self}=\textbf{CS-NPP}(\mathbf{F}_{input})$\;

\emph{Step2}.Embedding feature as sequence: 
$\mathbf{E}_{prior},\mathbf{E}_{self}=\textbf{HPE}(\mathbf{F}_{prior},\mathbf{F}_{self})$

\emph{Step3}.Feature reconstruction: 
$\mathbf{E}_{\mathrm{\mathcal{X}}}^{(0)}= \mathbf{E}_{prior},\mathbf{E}_{self},\mathcal{X}\in \left \{\mathrm{prior} ,\mathrm{self } \right \}$

\For{$l=1,...,{M}$}
{

$\mathbf{R}_{\mathcal{X}}^{(l)},\mathbf{E}_{\mathrm{S}}^{(l)}=\textbf{SACE}^{(l)}( \mathbf{E}_{\mathrm{\mathcal{X}}}^{(l-1)},\mathbf{E}_{\mathrm{S}}^{(l-1)})$\;

$\mathbf{E}_{\mathcal{X}}^{(l)},\mathbf{E}_{\mathrm{S}}^{(l)}=\textbf{CSCD}^{(l)}( \mathbf{R}_{\mathcal{X}}^{(l)},\mathbf{E}_{\mathrm{S}}^{(l)})$
	}
$\mathbf{E}_{prior}^\ast,\mathbf{E}_{self}^\ast=\mathbf{E}_{\mathcal{X}}^{(M)},\mathcal{X}\in \left \{\mathrm{prior} ,\mathrm{self } \right \}$

\emph{Step4}.Reverse mapping sequence into feature: 
$\mathbf{F}_{prior}^{\ast},\mathbf{F}_{self}^{\ast}=\textbf{HPE}^{-1}(\mathbf{E}_{prior}^\ast,\mathbf{E}_{self}^\ast)$

\textbf{Training Phase}: Calculate the loss using Eq. (6)

\textbf{Testing Phase}: Obtain anomaly score using Eq. (8)
\end{algorithm}
\begin{table*}
\caption{The quantitative results (image/pixel-level AUROC) of various methods on the MVTec AD dataset}
\setlength{\tabcolsep}{3pt}
\centering
\begin{tabular}{p{160mm}}
$\includegraphics[width=160mm]{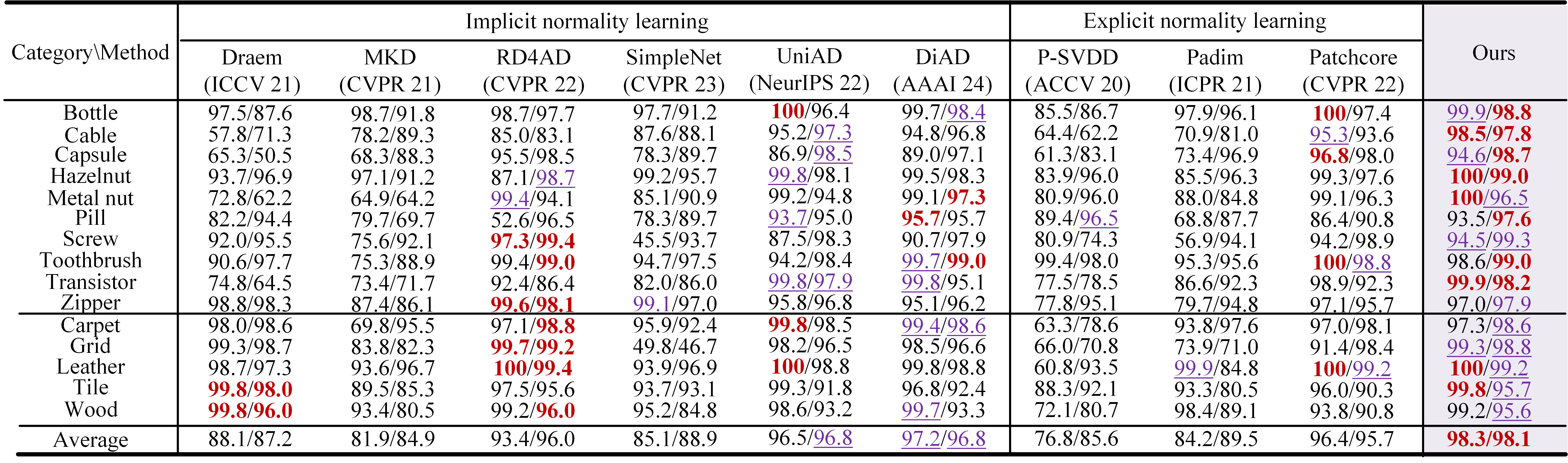}$
\end{tabular}
\begin{tablenotes}
        \footnotesize
        \item[1]The best outcomes are highlighted in bold red font, while the second-best results are indicated in blue font with underlining.
\end{tablenotes}
\label{table_mvtec}
\end{table*}

\begin{table*}
\caption{The quantitative results (logical/structural AUROC) of various methods on the MVTec LOCO dataset}
\setlength{\tabcolsep}{3pt}
\centering
\begin{tabular}{p{160mm}}
$\includegraphics[width=160mm]{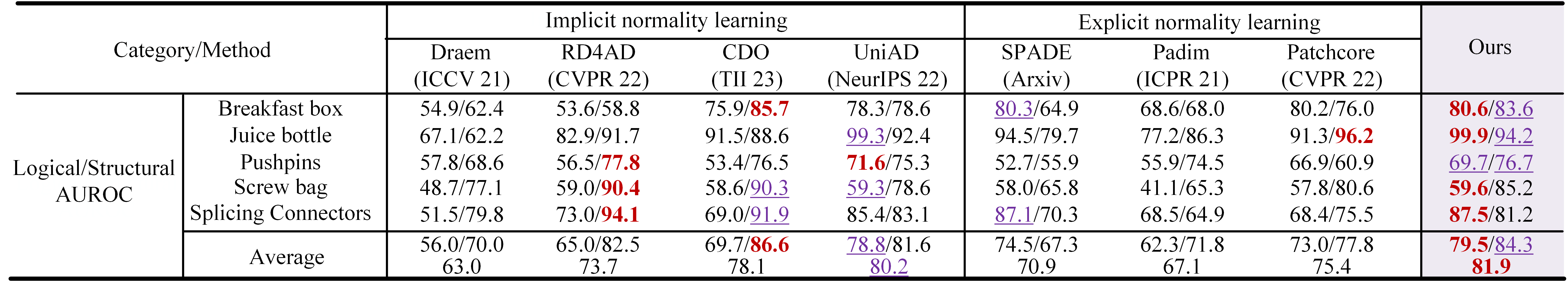}$
\end{tabular}
\label{table_loco}

\end{table*}

\section{Experimental verification}

This section offers empirical validation of the proposed PNPT, encompassing comparative analyses across various benchmark datasets, model ablation experiments, and real-world industrial applications.

\subsection{Experiments Setup}
\subsubsection{Dataset Descriptions}
This study rigorously evaluates the efficacy of the proposed PNPT across three benchmark datasets: the MVTec AD dataset~\cite{r3}, the MVTec LOCO AD dataset~\cite{r31}, and the BTAD dataset~\cite{r32}. The MVTec AD dataset, a widely recognized evaluation platform, comprises 3629 normal training images, 467 normal test images, and 1258 abnormal test images. The MVTec LOCO dataset, dedicated to logical anomaly detection, features five distinct categories. Additionally, the BTAD dataset, with 2540 images distributed across three categories, further diversifies the evaluation scenario. Furthermore, to assess the model's applicability in real industrial settings, we have also collected a private real-world dataset for button inspection.

\subsubsection{Implementation Details}
All images underwent resizing to dimensions of $256\times256$ and normalization using the mean and standard deviation derived from the ImageNet dataset. A wide-resnet50~\cite{r32} model was employed as the feature extractor. The PNPT architecture included $M=4$ pairs of SACE and CSCD. The hidden dimension $C$ was set at 760, the number of semantic tokens $N$ was 40, and the number of heads was configured to be eight. The PNPT model was trained from scratch for 300 epochs on only normal samples, employing the AdamW optimizer with a learning rate of 1e-4 and a batch size of eight.

It is noteworthy that the majority of existing methods originally adopted the one-model-one-class paradigm. This study replicates methods that have not previously been implemented in the multi-class setting in existing research to ensure fair comparisons. All experiments were conducted on a computing system equipped with Xeon(R) Gold 6226R CPUs@2.90 GHz, complemented by two NVIDIA A100 GPUs with 40GB of memory.

\subsubsection{Evaluation Metrics}

The performance of the proposed PNPT is evaluated using standard metrics, specifically the Area Under the Receiver Operating Characteristic Curve (AUROC). Image-level AUROC assesses the anomaly detection capacity, while pixel-level AUROC is employed to evaluate anomaly localization performance.

\subsection{Quantitative comparative experiment}

\subsubsection{MVTec AD}

Firstly, a comparative analysis was conducted on the MVTec AD dataset, comparing our proposed method against several representative approaches. The array of compared methods encompasses both implicit normality learning methods such as Draem~\cite{r4}, MKD~\cite{r11}, RD4AD~\cite{r16}, SimpleNet~\cite{r30}, and the recent methods UniAD~\cite{r14} and DiAD~\cite{r34}, as well as explicit normality learning methods P-SVDD~\cite{r20}, Padim~\cite{r19}, and Patchcore~\cite{r5}. 
It is noteworthy that UniAD and DiAD have been specifically tailored for multi-category industrial anomaly detection. UniAD serves as the baseline model in this context, while DiAD represents the latest state-of-the-art model. The outcomes of this evaluation are meticulously detailed in Table~\ref{table_mvtec}. It is evident from the results that the proposed PNPT outperforms all competitors at both the pixel and image levels. Notably, PNPT exhibits improvements of +1.8 in image-level AUROC and +1.3 in pixel-level AUROC when compared to the baseline model UniAD in the realm of multi-class anomaly detection. Furthermore, our method also surpasses the latest model, DiAD, demonstrating enhancements in image and pixel-level AUROC by +1.1 and +1.3, respectively.

While attaining the most optimal overall outcomes, our approach consistently maintains stable and outstanding performance across diverse categories. Evidently, PNPT exhibits optimal or near-optimal performance across all 15 categories. Particularly noteworthy is the substantial improvement in image AUROC for the "capsule" and "screw" categories, which increased by +7.7 and +7.0, respectively, in contrast to UniAD.

\subsubsection{MVTec LOCO} As a recently introduced dataset, MVTec LOCO presents greater challenges compared to MVTec AD, attributed to the inclusion of logical defects. Furthermore, the level of difficulty is further heightened within the unified paradigm of multi-class training. We have also chosen a selection of representative methods encompassing both implicit learning approaches, namely RD4AD~\cite{r16}, CDO~\cite{r25}, Draem~\cite{r4}, and UniAD~\cite{r14}, and explicit learning methods such as SPADE~\cite{r18}, Padim~\cite{r19}, and Patchcore~\cite{r5}. The quantitative comparison results pertaining to their evaluations on logical defects and structural defects in MVTec LOCO are presented in Table~\ref{table_loco}.

The outcomes indicate that PNPT consistently attains a superior overall performance of 81.9. In contrast to the sub-optimal UniAD, noteworthy enhancements of +1.7 were realized. Furthermore, the improvement in performance in detecting logical defects is more pronounced compared to other alternative models. It is worth noting that compared with RD4AD, CDO, and Patchcore, our performance in detecting logic defects shows substantial improvements of +14.6, +9.8, and +6.5 respectively. However, even with these advances, the performance of existing methods, including the proposed PNPT, remains unsatisfactory in the Screw bag category. This is attributed to the challenges posed by the complexity of this category.

\begin{table}
\caption{The image/pixel-level AUROC results on the BTAD dataset}
\label{table}
\setlength{\tabcolsep}{3pt}
\centering
\begin{tabular}{p{85mm}}
${\includegraphics[width=85mm]{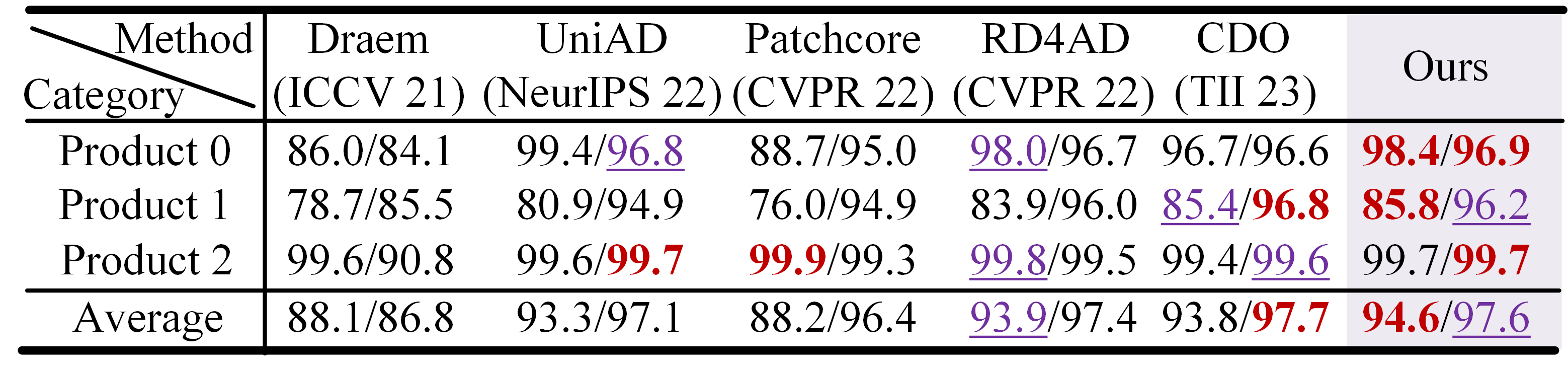}}$
\end{tabular}
\label{table_btad}
\end{table}

\subsubsection{BTAD}
Moreover, additional comparative experiments were conducted on the intricate texture dataset, BTAD. The AUROC results at both the image and pixel levels are cataloged in Table~\ref{table_btad}. The findings illustrate that when compared to existing advanced methods Draem~\cite{r4}, and UniAD~\cite{r14}, Patchcore~\cite{r5}, RD4AD~\cite{r16} and CDO~\cite{r25}, PNPT consistently outperforms in terms of performance.

Overall, PNPT has showcased state-of-the-art performance within the multi-class unified scenario across three publicly benchmark datasets, highlighting the advancement of PNPT.

\begin{figure*}[t]
\centerline{\includegraphics[width=150mm]{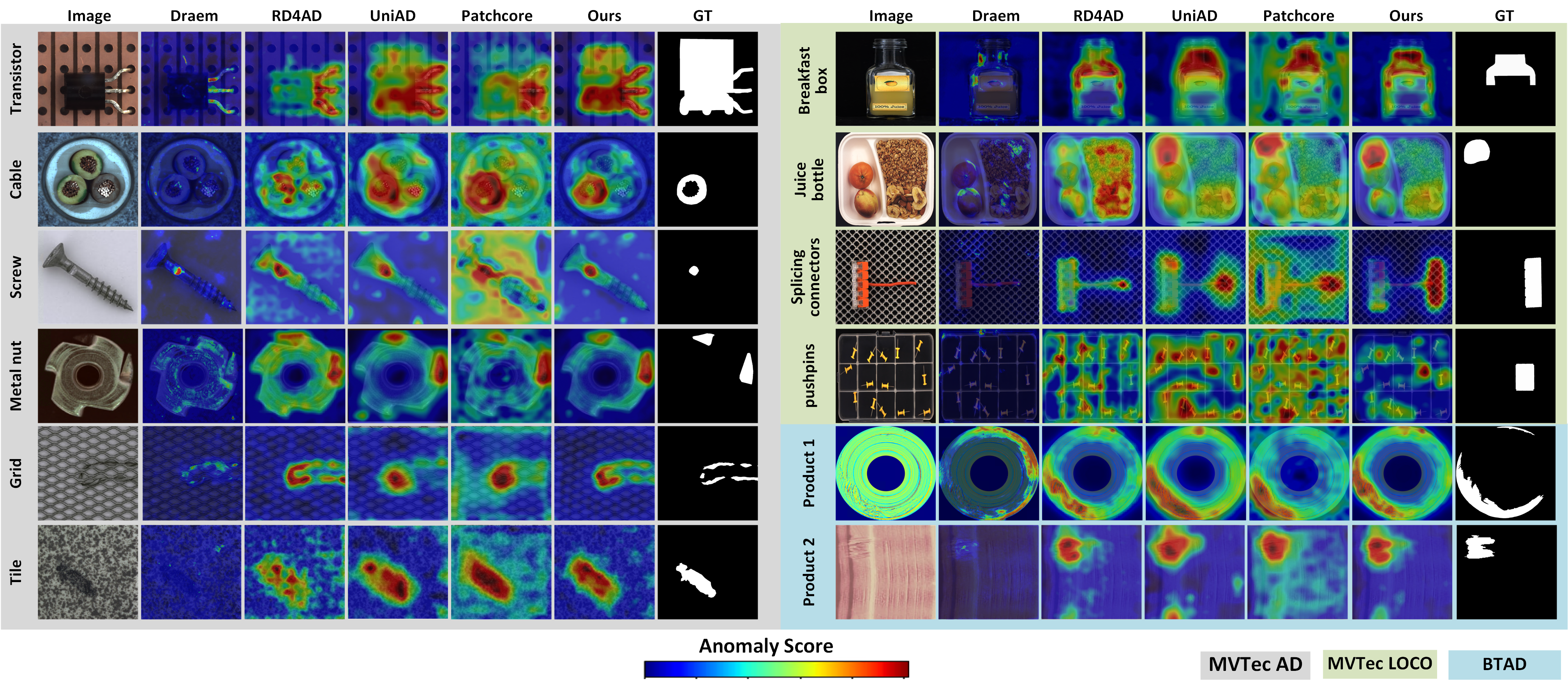}}
\caption[width=150mm]{
The qualitative detection outcomes of the proposed model and the comparative methods on the MVTec AD, MVTec LOCO, and BTAD benchmarks. In contrast to alternative approaches, our method exhibits superior detection accuracy across various samples and defect types.}
\label{fig6}
\end{figure*}

\subsubsection{Computational complexity analysis}

Considering the applications for industrial scenarios, assessing the computational complexity of the model is crucial. This study conducted comparisons between our proposed PNPT and several established methods using the MVTec AD dataset, outlined in Table~\ref{table_cost}. The experimental results indicate the following: \textbf{\emph{1}})~Explicit methods like Padim and Patchcore can lead to substantial training memory consumption in multi-category scenarios due to their expansive memory banks. \textbf{\emph{2}})~Despite a slight increase in computational complexity compared to other CNN-based implicit methods due to the vision Transformer-based structure, PNPT still maintains comparable Floating Point Operations (FLOPs) and inference time.

In summary, PNPT has moderate computational demands, necessitating around 10 hours and 5.7G GPU memory to complete training for 15 categories in the MVTec AD dataset, and only 53.7 milliseconds(ms) per image during testing.

\begin{table}
\caption{Comparative results of computational costs}
\label{table}
\setlength{\tabcolsep}{3pt}
\centering
\begin{tabular}{p{80mm}}
${\includegraphics[width=80mm]{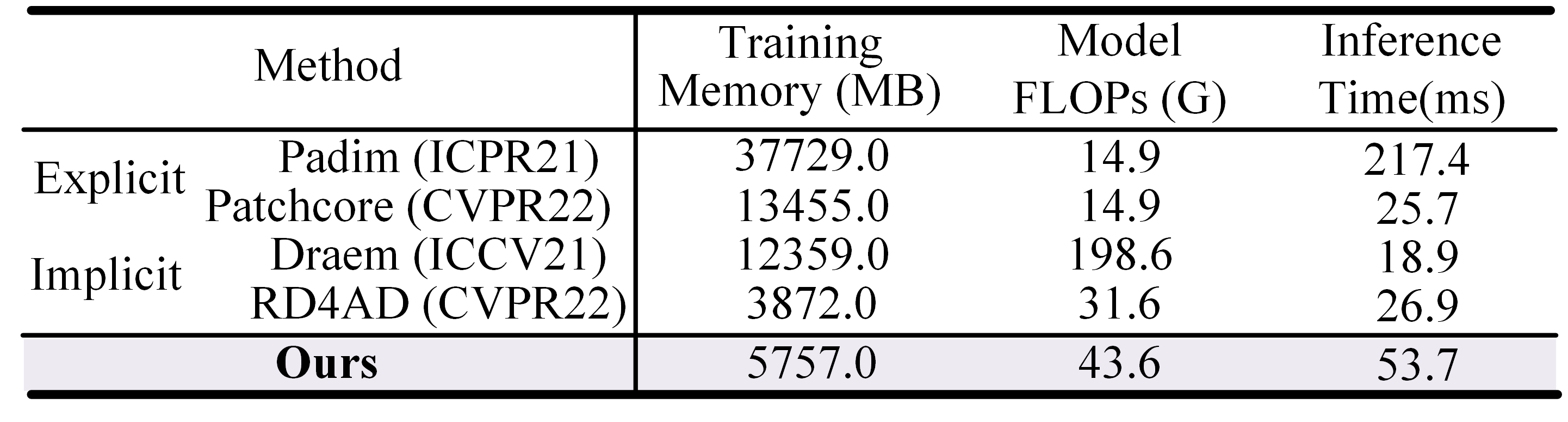}}$
\end{tabular}
\label{table_cost}
\end{table}

\subsection{Qualitative comparison results}

To assess our method's effectiveness, we conducted qualitative comparative experiments. Fig.~\ref{fig6} displays anomaly localization outcomes from various methods on representative examples from MVTec AD, MVTec LOCO, and BTAD datasets. The analysis highlights key factors contributing to our PNPT approach's outstanding performance: \textbf{\emph{1}})~Our approach demonstrates robustness when applied to multi-category anomaly detection tasks. In contrast, the implicit method DRAEM performs poorly in this intricate scenario. \textbf{\emph{2}})~Owing to the hierarchical feature representation, our method is capable of producing more precise anomaly localization maps with reduced edge uncertainty and increased compactness. In contrast, the localization maps obtained by UniAD and Patchcore in the low-resolution feature space tend to be coarser. \textbf{\emph{3}})~Our method exhibits excellent detection capability for logical anomalies. In contrast, comparative methods are better suited for detecting structural defects, such as gray strokes on tiles, as they generally focus on extracting local semantics, exemplified by memory bank entries in Patchcore that store local patterns. However, when it comes to higher-level semantic anomalies like missing connectors in splicing connectors and absent pushpins, these comparative methods struggle to achieve accurate localization results. Conversely, our approach exhibits superior high-level semantic acquisition capabilities, enabling it to effectively detect such defects.

\begin{figure}[t]
\centerline{\includegraphics[width=80mm]{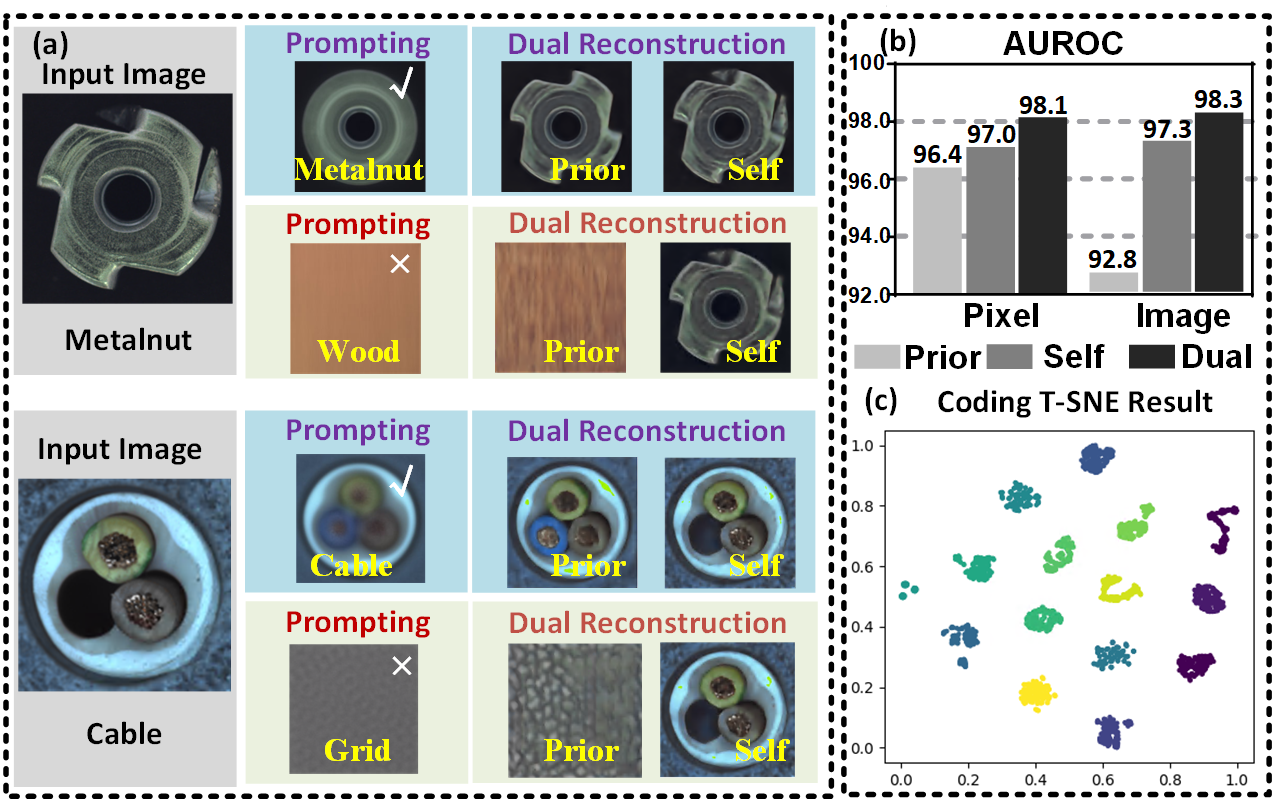}}
\caption[width=80mm]{
Visualization effect of the prior normality prompting. (a) The dual-reconstruction results of the model under different prior normality prompting.  (b) Quantitative AUROC results of the dual branches. (c) The latent distribution of codings across 15 categories in MVTec AD.
}
\label{fig7}
\end{figure}

\subsection{Ablation experiment}

\begin{figure*}[t]
\centerline{\includegraphics[width=170mm]{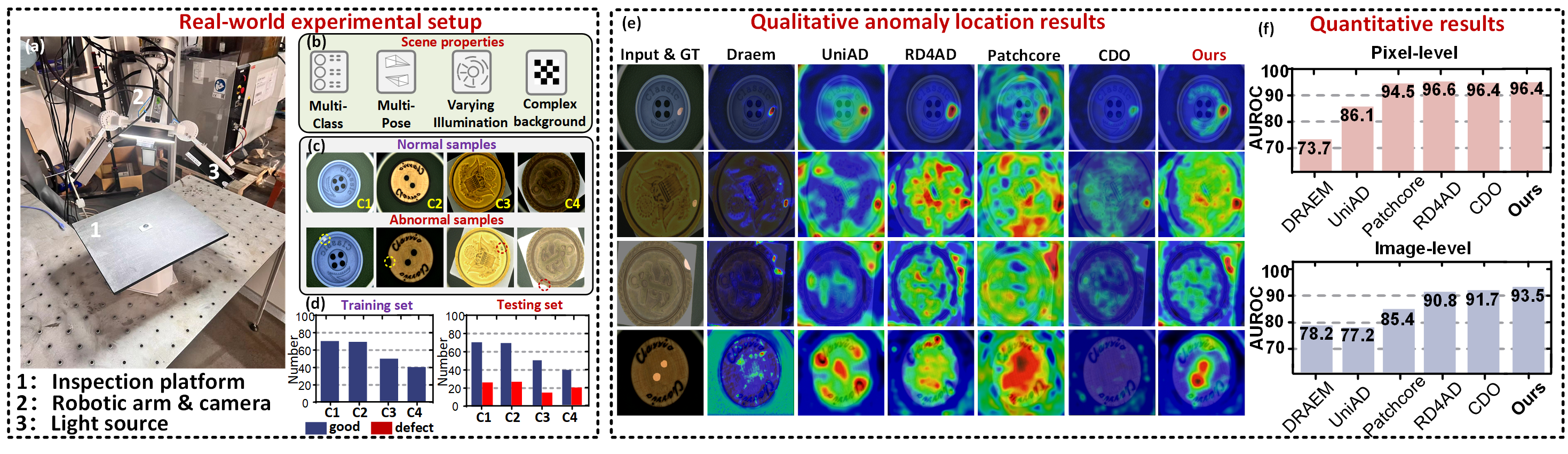}}
\caption[width=170mm]{
Experimental verification results for real-world applications. (a) Automatic optical inspection equipment used in the experiment (b) Scene properties considered during data collection. (c) Sample instances of normal and abnormal buttons. (d) The specific configuration of the constructed data set.
(e)-(f) Comparison of qualitative and quantitative results of different methods.}
\label{fig8}
\end{figure*}
In this section, ablation experiments were conducted to evaluate each module's specific contribution to overall performance in the proposed methodology. Quantitative results are presented in Table~\ref{table_ablation}, following the format of the earlier comparative study. Among them, model Variant (A) acts as the baseline vision Transformer (ViT) model.

\begin{table}
\caption{AUROC for different variants of ablation experiments}
\label{table}
\setlength{\tabcolsep}{3pt}
\centering
\begin{tabular}{p{80mm}}
${\includegraphics[width=80mm]{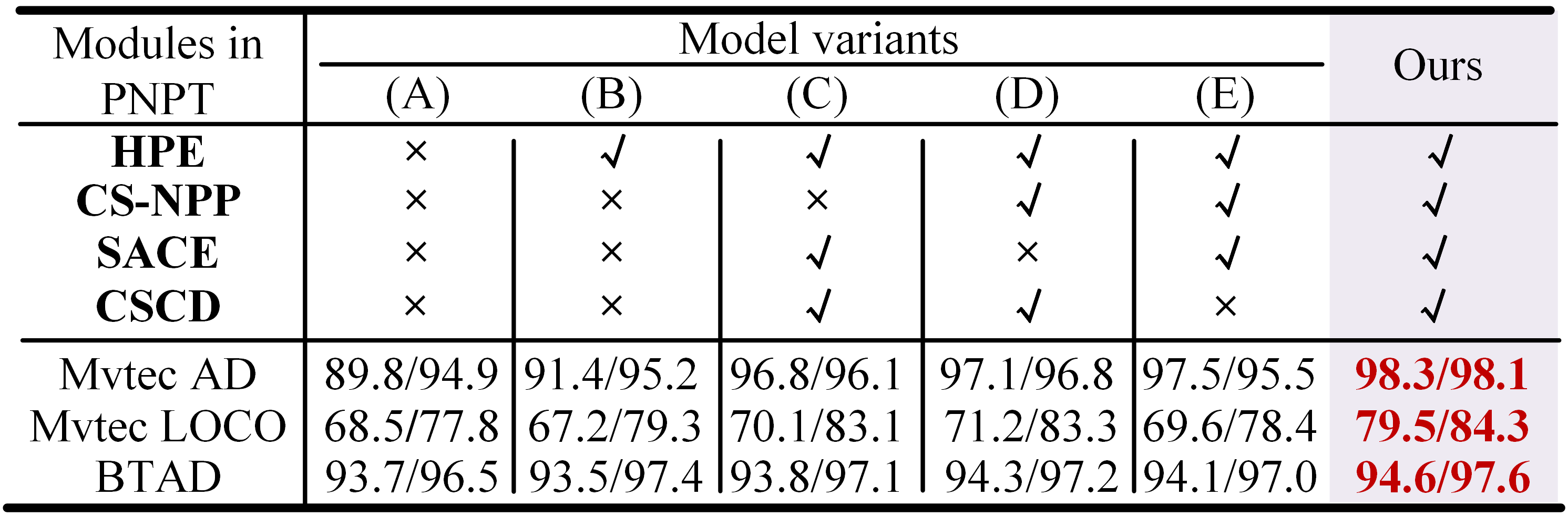}}$
\end{tabular}
\label{table_ablation}
\end{table}

\subsubsection{HPE} HPE is employed to acquire multi-scale feature information. Upon removing HPE, the reconstruction model only adopts the last layer of information from the pre-trained CNN. A comparative analysis of model variants (A) and (B) reveals the performance enhancement attributable to HPE.

\subsubsection{CS-NPP} CS-NPP provides category-specific normal prompt information, serving as the fundamental rationale behind this methodology. In the absence of CS-NPP, the model regresses into a single-stream architecture. The outcomes from model variant (C) demonstrate a noticeable decline in performance on both datasets, particularly evident in addressing logical defects within MVTec LOCO. A more intuitive visualization analysis is presented in Fig.~\ref{fig7}.

As depicted in Fig.~\ref{fig7} (a), the prior reconstruction is effectively guided by prior information, leading to the restoration of anomalies into normal semantics when given the correct prompting. To further assess the impact of prior normality prompting, this study compulsively fixes the category of the employed prior prompting information. The results show that the model's prior reconstruction aligns with the provided prompting normality of the error category, while the self-reconstruction remains relatively stable. As illustrated in Fig.~\ref{fig7} (b), optimal performance is achieved by fusing the reconstruction outcomes from both the prior and self-attribute branches.

Furthermore, to evaluate the retrieval stability of CS-NPP, as shown in Fig.~\ref{fig7} (c), this study conducted T-SNE dimensionality reduction visualization on the codings of the 15 categories of samples in MVTec AD. The observed distribution demonstrates high discriminative capability, and our experimental findings indicate the stability of correct retrieval.

\subsubsection{SACE} SACE is applied to extract and fuse the contextual semantics of sample self-properties and the prior normality. The exclusion of SACE entails discarding the semantic tokens $\mathbf{E}_{\mathcal{S}}$, utilizing only the patch tokens for alignment and fusion, resulting in the creation of variant (D). A comparative analysis reveals a performance reduction in variant (D) in comparison to the complete model, particularly noticeable in addressing logical defects within MVTec LOCO that necessitate global semantic understanding. This is attributed to the fact that semantic tokens facilitate contextual semantic extraction through the establishment of long-distance semantic dependencies. The elimination of these tokens results in the model losing its capacity for context extraction.

\subsubsection{CSCD} CSCD is employed to decode the fused contextual semantics obtained through SACE. The omission of CSCD implies the exclusion of contextual semantic tokens during decoding, resulting in the reconstruction of the model solely using SACE-encoded patch tokens. Model variant (E) demonstrates a notable decline in performance, as the lack of contextual semantic tokens following alignment and fusion hinders interaction between the prior branch and the self-attribute branch. Consequently, the prior branch lacks information about the sample's self-attributes, preventing it from recovering a normal template closely resembling the sample. Additionally, the self branch also adopts identical mapping shortcut learning due to the absence of prior normal semantics.

\subsection{Real-world applications}

To further assess the applicability and generalization of the proposed PNPT in real industrial scenarios, we applied it to a real-world task involving the detection of button defects.

The automatic optical inspection equipment employed in our study is depicted in Fig.~\ref{fig8} (a). It features a robotic arm equipped with a light source and camera, utilized for capturing images of buttons. To assess the detection capabilities in complex industrial scenarios, we intentionally introduced multiple interference attributes during sample collection as shown in Fig.~\ref{fig8} (b). Firstly, we collected multiple classes (four classes, \textbf{C1}-\textbf{C4}) of buttons to evaluate the model's multi-category detection capabilities. Secondly, in contrast to existing benchmarks like MVTec AD, MVTec LOCO, and BTAD, our collected images encompass buttons under various poses, incorporate changing lighting conditions, and introduce background interference. These intentional variations significantly enhance the complexity of the dataset, rendering it more representative of real industrial scenes. Fig.~\ref{fig8} (c) displays a selection of collected normal and abnormal samples. Evidently, these samples showcase substantial variability. The constructed button dataset configuration, as depicted in Fig.~\ref{fig8} (d), comprises a total of 230 normal samples for the training set, and 230 normal samples along with 87 abnormal samples for the testing set.

The qualitative anomaly localization comparison results of various methods are depicted in Fig.~\ref{fig8} (e). It is evident that across different types of buttons and diverse defects, our PNPT method consistently achieves the most accurate anomaly localization results. In the second row, concerning tiny defects, UniAD, RD4AD, and Patchcore exhibit abnormal response noise in the normal area. In the third row, where background interference is introduced, UniAD and Patchcore encounter challenges. Notably, for the most challenging type of logical anomaly—missing holes—depicted in the fourth row, only our method demonstrates the capability to detect such defects.

\begin{table}
\caption{The pixel/image-level AUROC results on the button dataset}
\label{table}
\setlength{\tabcolsep}{3pt}
\centering
\begin{tabular}{p{80mm}}
${\includegraphics[width=80mm]{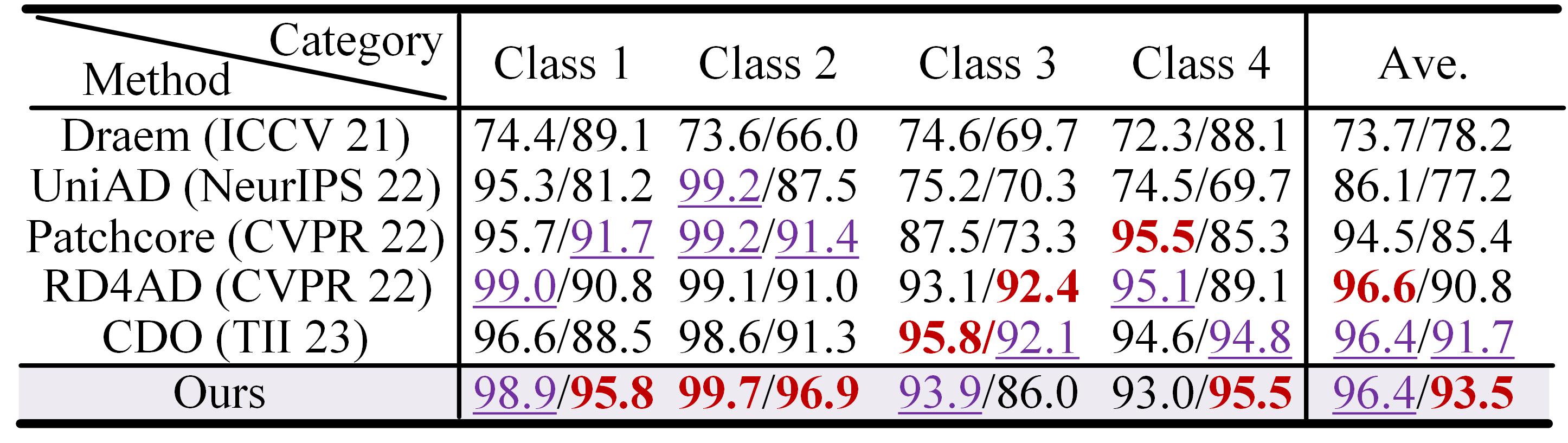}}$
\end{tabular}
\label{table_button}
\end{table}

In Fig.~\ref{fig8} (f) and Table~\ref{table_button}, comprehensive and class-by-class quantitative comparisons of image and pixel-level AUROC are presented. Remarkably, our method demonstrates excellent overall performance and stable detection across four categories, achieving optimal or sub-optimal results in each category.

\subsection{Discussion and limitations}

While our experiments sufficiently validate the efficacy of the proposed PNPT, we acknowledge certain limitations. Notably, challenges arise in detecting product defects with intricate logical constraints, as seen in the case of the Screw Bag. The disordered arrangement and strong semantic logical constraints pose difficulties for existing single-modal vision methods, resulting in unsatisfactory outcomes. In future investigations, we will endeavor to explore the implementation of the Large-scale visual-linguistic model\cite{r35} to effectively address these anomalies. Additionally, the button dataset used in the real-world verification is constrained by the limited number of abnormal samples. We will continue to collect defect samples to expand this dataset and enhance our contribution.
\section{Conclusion}

In conclusion, this paper has introduced the PNPT framework, a unified model specifically crafted for multi-class industrial anomaly detection. With the semantic alignment of prior normality prompting and sample self-attributes, PNPT stands out as a robust and identical-mapping-resistant reconstruction model. Comprehensive experimentation across three widely recognized benchmark datasets affirming the superiority of PNPT in detecting multi-class anomalies simultaneously. In future research endeavors, we intend to apply this framework to a broader spectrum of industrial scenarios.

\bibliographystyle{ieeetr} 
\bibliography{reference}

\begin{IEEEbiography}[{\includegraphics[width=1in,height=1.25in, clip,keepaspectratio]{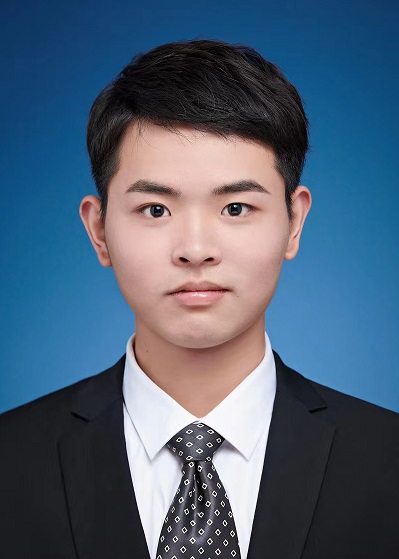}}]{Haiming Yao} (Graduate Student Member, IEEE) received a B.S. degree(Hons.) from the School of Mechanical Science and Engineering, Huazhong University of Science and Technology, Wuhan, China, in 2022. He is pursuing a Ph.D. degree with the Department of Precision Instrument, Tsinghua University, Beijing, China.

His research interests include visual anomaly detection, deep learning, visual understanding, and artificial intelligence for science.
\end{IEEEbiography}
\vspace*{-5mm}
\begin{IEEEbiography}[{\includegraphics[width=1in,height=1.25in,clip,keepaspectratio]{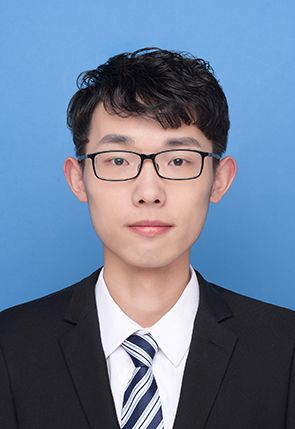}}]{Yunkang Cao}(Graduate Student Member, IEEE) received the B.S. degree in mechanical design,
 manufacturing and automation from the Huazhong
 University of Science and Technology, Wuhan,
 China, in 2020, where he is currently pursuing the
 Ph.D. degree in mechanical engineering.
 
 His current research interests include industrial
 foundation models and their real-world applications,
 anomaly detection, and computer vision.
\end{IEEEbiography}

\begin{IEEEbiography}[{\includegraphics[width=1in,height=1.25in,clip,keepaspectratio]{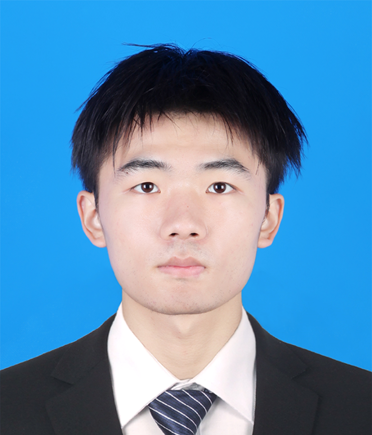}}]{Wei Luo}(Student Member, IEEE) received a B.S. degree from the School of Mechanical Science and Engineering, Huazhong University of Science and Technology, Wuhan, China, in 2023. He is pursuing a Ph.D. degree with the Department of Precision Instrument, Tsinghua University, Beijing, China.

His research interests include deep learning, anomaly detection and machine vision.
\end{IEEEbiography}

\begin{IEEEbiography}[{\includegraphics[width=1in,height=1.25in,clip,keepaspectratio]{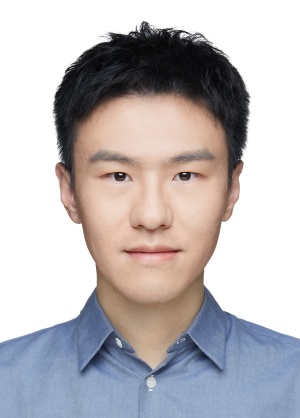}}]{Weihang Zhang} received the B.S. degree from the School of Instrumentation Science and Opto-Electronics Engineering, Beihang University, Beijing, China, in 2015, and the Ph.D. degree from the Department of Precision Instrument, Tsinghua University, in 2020. He was a Postdoctoral Researcher with the Department of Precision Instrument, Tsinghua University, from 2020 to 2022. He is currently an Assistant Professor with the School of Medical Technology, Beijing Institute of Technology.

His research interests include image processing, pattern recognition, and intelligent computing. 

\end{IEEEbiography}

\begin{IEEEbiography}[{\includegraphics[width=1in,height=1.25in,clip,keepaspectratio]{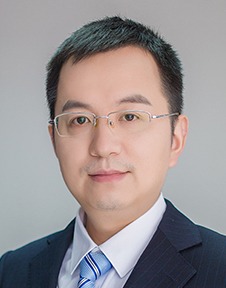}}]{Wenyong Yu}(Senior Member, IEEE) received an M.S. degree and a Ph.D. degree from Huazhong University of Science and Technology, Wuhan, China, in 1999 and 2004, respectively. He is currently an Associate Professor with the School of Mechanical Science and Engineering, Huazhong University of Science and Technology. 

His research interests include machine vision, intelligent control, and image processing.
\end{IEEEbiography}

\begin{IEEEbiography}[{\includegraphics[width=1in,height=1.25in,clip,keepaspectratio]{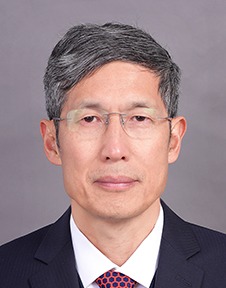}}]{Weiming Shen} (Fellow, IEEE) received the B.E. and M.S. degrees in mechanical engineering from Northern Jiaotong University, Beijing, China, in 1983 and 1986, respectively, and the Ph.D. degree in system control from the University of Technology of Compiègne, Compiègne, France, in 1996.

 He is currently a Professor with the Huazhong University of Science and Technology (HUST), Wuhan, China, and an Adjunct Professor with the University of Western Ontario, London, ON,
 Canada. Before joining HUST in 2019, he was a Principal Research Officer with the National Research Council Canada. His work has been cited more than 20000 times with a h-index of 70. He
 authored or co-authored several books and more than 600 articles in scientific journals and international conferences in related areas. His research interests include agent-based collaboration technologies and applications, collaborative intelligent manufacturing, Internet of Things, and big data analytics.
 
 Prof. Shen is a Fellow of the Canadian Academy of Engineering and the Engineering Institute of Canada.

\end{IEEEbiography}

\end{document}